# Image Retrieval Based on LBP Pyramidal Multiresolution using Reversible Watermarking

H. Ouahi*[1], K. Afdel*[2], M.Machkour*[3]

*Laboratory Computing Systems & Vision LabSiv
University Ibn Zohr of Agadir

Faculty of Science, BP 28/S, Agadir 80000
MOROCCO

*Abstract*— In the medical field, images are increasingly used to facilitate diagnosis of diseases. These images are stored in multimedia databases accompanied by doctor's prescriptions and other information related to patients. Search for medical images has become for clinical applications an essential tool to bring effective aid in diagnosis. Content Based Image Retrieval (CBIR) is one of the possible solutions to effectively manage these databases. Our contribution is to define a relevant descriptor to retrieve images based on multiresolution analysis of texture using Local Binary Pattern LBP. This descriptor once calculated and information's relating to the patient; will be placed in the image using the technique of reversible watermarking. Thereby, the image, descriptor of its contents, the BFILE locator and patient-related information become a single entity, so even the administrator cannot have access to the patient private data.

## I. INTRODUCTION

With the appearance of image acquisition devices, the medical field produces a large number of images. However, the fast and accurate access to image databases requires the use of efficient indexing algorithms. In the literature [1,2,3], we find several image indexing and retrieval systems, the first systems are based on the text, using a technique that consists in describing the visual content in text form (keywords), those keywords are used as index. The advantage of this technique is that it allows the use of SQL. Its disadvantage, however, is that it requires a large amount of manual processing, and the text used may not describe better the content of the image. To solve these problems the CBIR systems have emerged, these systems use the visual content of the image to construct the index. The descriptor choice, describing the image, is a key element for the search and retrieval of image, especially for mammography images that don't have evident structure, and the only information that can be used to characterize them is related to the pixels greyscale values. For this reason, we have decided to use the descriptor based on texture analysis using LBP. Several image retrieval systems by content exists; on the one hand, there are commercial systems such as IBM's QBIC in 1995, Image Finder, Excalibur, and other experimental systems as Blobworld, photobook, Ikona, Viper. In all these database systems, images and data corresponding patients are separated. Thus, the personal data of patients can be consulted by the administrator of the database that constitutes a serious obstacle to the privacy of patients which needs to be protected. On the other hand, given that the images are stored outside the database, therefore, they do not have the security procedures of the database. The link in the database between images, descriptors and patient information may be lost. To remedy these problems, we will use the watermarking technique to integrate the image descriptor describing its contents, BFILE locator and patients' information. The watermarking technique should provide a large data integration capability and must also be reversible to preserve the contents of the medical image which must not be modified.

This paper is organized into three main sections. Section II presents the overall architecture of an indexing system image by content; Section III describes the architecture of our system based on the watermarking, while Section IV is devoted to present the experimental results on medical images. The conclusion, on the other hand, is stated in the last section.

## II. THE GLOBAL ARCHITECTURE OF AN IMAGE RETRIEVAL SYSTEM BY CONTENT

In multimedia database, indexing is used to describe the images by their content using descriptors consist of low-level parameters related to texture, shape and / or color; for each image corresponds one or more feature vectors forming the index of the image. However, an image indexing system by content consists of two major steps:

- The first in **offline mode** (indexing system) is to calculate a signature from the descriptors of the image in a way to permit a quickly access data, this descriptor is stored in the images database; in our case, it is embedded directly into the image using the reversible watermarking technique.

- The second **online mode** (search system) is to each query image, the user selects an image using a graphical interface, the index of the query image are calculated and compared to those of the database image, and finally the system selects the most similar images.

Thus the two sub-systems have in common the following two treatments:
• The feature extraction descriptors images
• The generation of index from the descriptors.

Fig1 shows the architecture of search by content system. [4][5]

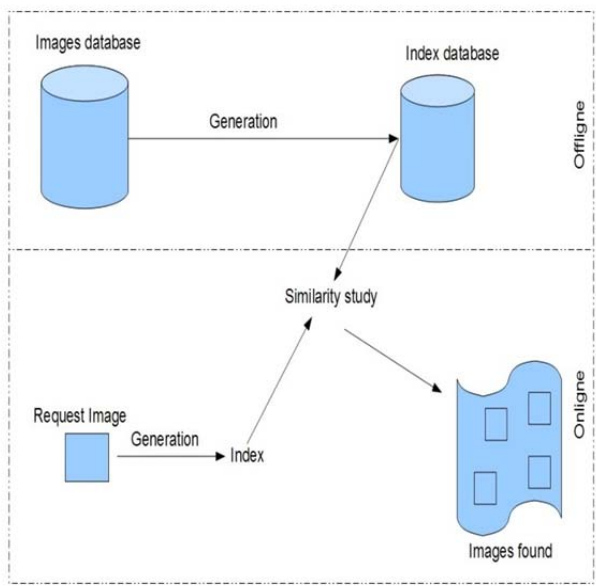

Fig. 1  CBIR system architecture

## III. THE APPROACH IMAGE SEARCH BY REVERSIBLE WATERMARKING

### A. Local binary pattern descriptor LBP

There are several content descriptors of the image. These descriptors are calculated by exploiting the content of the image. For medical images in general and mammography images especially, texture and color analysis lead to relevant and robust descriptors. LBP is a powerful feature that is used to describe textures for the detection and tracking of objects in the field of computer vision. The formula of this descriptor is given by:

$$LBP_{x,a} = \sum_{x=0}^{x-1} sign(vx - va) 2^x$$

x Є [0,8]
sign(p) = 1 si p>=0 ou 0 si p<0

Fig.2 shows the LBP neighborhood with distance and the number of neighbors.

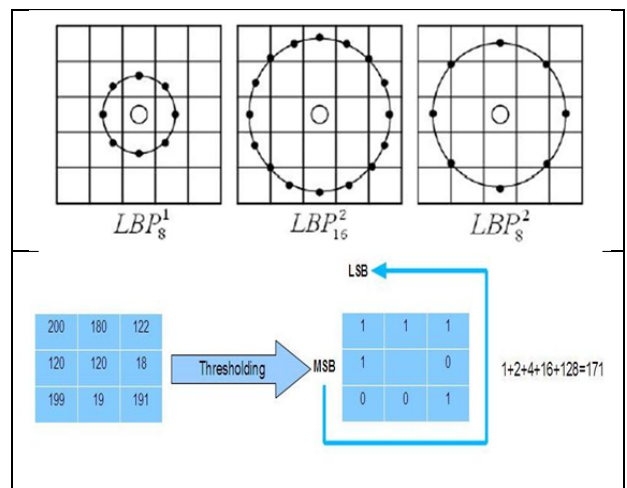

Fig. 2  LBP neighborhood

Each pixel of the image is encoded by thresholding its 3 * 3 neighborhood by its value and assuming that the result is a binary number as shown in the example of Fig. 2 [6][7].

### B. Gaussian pyramidal multiresolution

The techniques using multiresolution are very important for two reasons: First, they reduce the number of operations and the time required to process an image. Second, they allow for easy search and retrieval objects with different size in the image, or just to accommodate different resolutions of the images in the images database. The pyramidal multiresolution analysis used is described by the pyramid model introduced by Tanimoto and Pavlidis [8] Fig.3.

A pyramid I is a sequence of images $I_h$, each image corresponding to a version of the original image $I_0$ for reduced resolution.

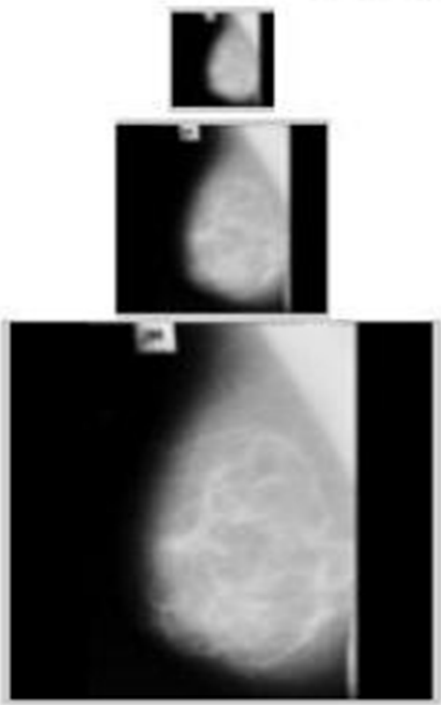

Fig. 3  Pyramidal multiresolution I0, I0/2, I0/4

In our approach, we use the accumulated LBP histograms of three levels resolution, which is given by the following definition:

The accumulation of two LBP histograms G and I is given by the following mathematical formula:

CHLBPL(G,I)= $\sum_{i=0}^{N}(g_i + h_i)$

Below, the Fig.4 shows the cumulative histogram LBP of levels histograms $L_0$, $L_1$ and $L_2$:

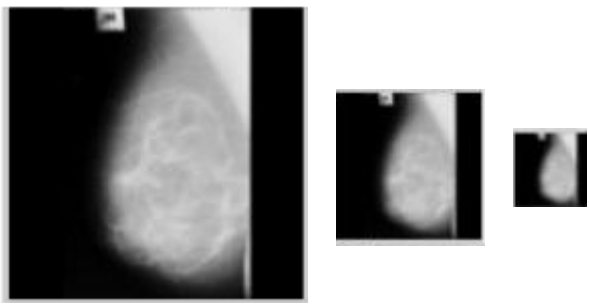

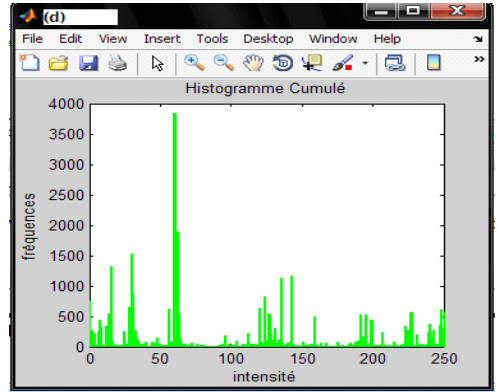

Fig. 4  cumulative histogram LBP

*C. Reversible watermarking*

There are several watermarking techniques in the literature, we choose the reversible watermarking because the content of the medical image should in no case be modified. We used the fragile reversible watermarking method based on the difference expansion; the watermark is embedded into LSB bits of the difference value.

If we take a pair of greyscale value of pixel (x, y); x,y ∈ Z, (0 <=x,y<= 255),

The average value is given by: $l = \lfloor \frac{x+y}{2} \rfloor$ and the difference is given by: $h = x - y$

The symbol ⌊.⌋ represents the integer value less than or equal. To recover the original value of the pixel, we simply need to reverse the above equations:

$$x = l + \lfloor \frac{h+1}{2} \rfloor, \qquad y = l - \lfloor \frac{h}{2} \rfloor$$

Knowing the values of gray levels (ng) of the pixels must belong to the interval [0,255], it is necessary that $\left(0 \leq l + \lfloor \frac{h+1}{2} \rfloor \leq 255\right), \left(0 \leq l - \lfloor \frac{h}{2} \rfloor \leq 255\right)$ these two inequalities are equivalent to $|h| \leq \min(2(255 - l), 2l + 1)$.

During the mark integration in the image, the pixels of the image are grouped into three zones according to the difference value of h: the extensible zone, changing zone and the unchanging zone. The data bits of the payload are included only in extensible and changeable zone [9]. The Location map bits integration is established to distinguish these three zones. [10]

*D. Our approach*
*1) Database watermarking:* The Fig.5 shows the image insertion procedure in the database

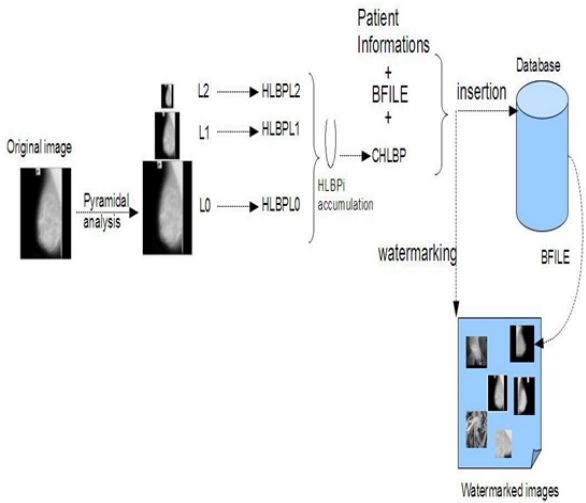

Fig. 5  LBP watermarking system image insertion

Images database watermarking algorithm:

- Original image decomposition on three pyramid levels $L_0$, $L_1$ and $L_2$;
- Determining the LBP histogram of each level Li ;
- Calculation of accumulated three histograms HLBPi (i=0,1,2) ;
- Embed the CHLBP descriptor, the BFILE locator and the patient information(ID, name, birthday, diagnostic information) in the image using the reversible watermarking technique;
- Store the watermarked image in a file system directory, and his locator BFILE in the database.

  Thus, the embedded information's allows us to make multi-criteria Search, by using CHLBP descriptor or the patient ID which are inserted in the watermarked image; the embedded BFILE locator allows us to restore link between the watermarked images and the database in case of breakage.

*2) Image retrieval online:* Our system uses the image retrieval based on the reversible watermarking technique to preserve the original image. We calculated LBP histograms for the levels $L_0$, $L_1$ and $L_2$ of the Gaussian pyramidal multiresolution. Then, we performed the cumulative histograms for this three resolution levels $L_0$, $L_1$ and $L_2$ to build our descriptor.

The search and retrieval of image is described in the following Fig.6:

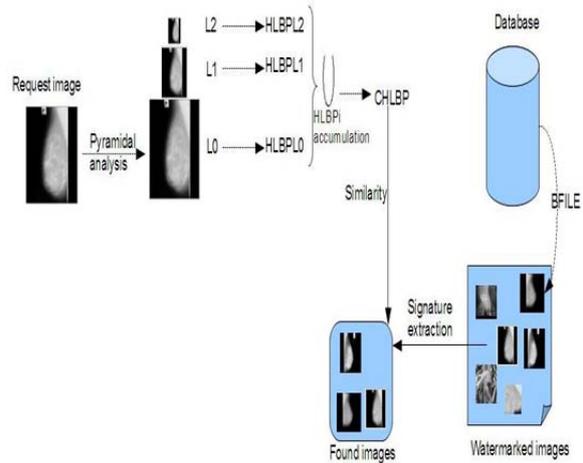

Fig. 6  LBP watermarking system image extracting

Image retrieval algorithm:

- Request image decomposition on three pyramid levels L0, L1 and L2;
- Determining the LBP histogram of each level Li ;
- Calculation of accumulated three histograms HLBPi (i=0,1,2) ;
- For the search and retrieval of an image, we perform the calculation of the Euclidean distance between the values of the cumulative histogram of the query image and those extracted from the watermarked image database and selected the images with a smallest Euclidean distance value.

IV. EXPERIMENTS AND RESULTS

The performance evaluation for CBIR systems is often done by the following two measures: recall and precision. [11]

**Recall:**
The recall is the ratio between the number of relevant images in the set of images found and the number of relevant images in the images database.

$$Rappel=|Ra| / |R|$$

**Precision:**
Precision is the ratio between the number of relevant images in the set of images found and the number of images found.

$$Precision=|R_a| / |A|$$

Where R is the set of relevant images in the images database used to evaluate (R is the set of images similar to the query), |R | is the number of relevant images in the

image database, A is the set of responses, |A| is the number of images in the set of responses and |Ra| is the number of relevant images in the set of responses.

Fig.7 shows the average precision and recall to three images of each of the three classes.

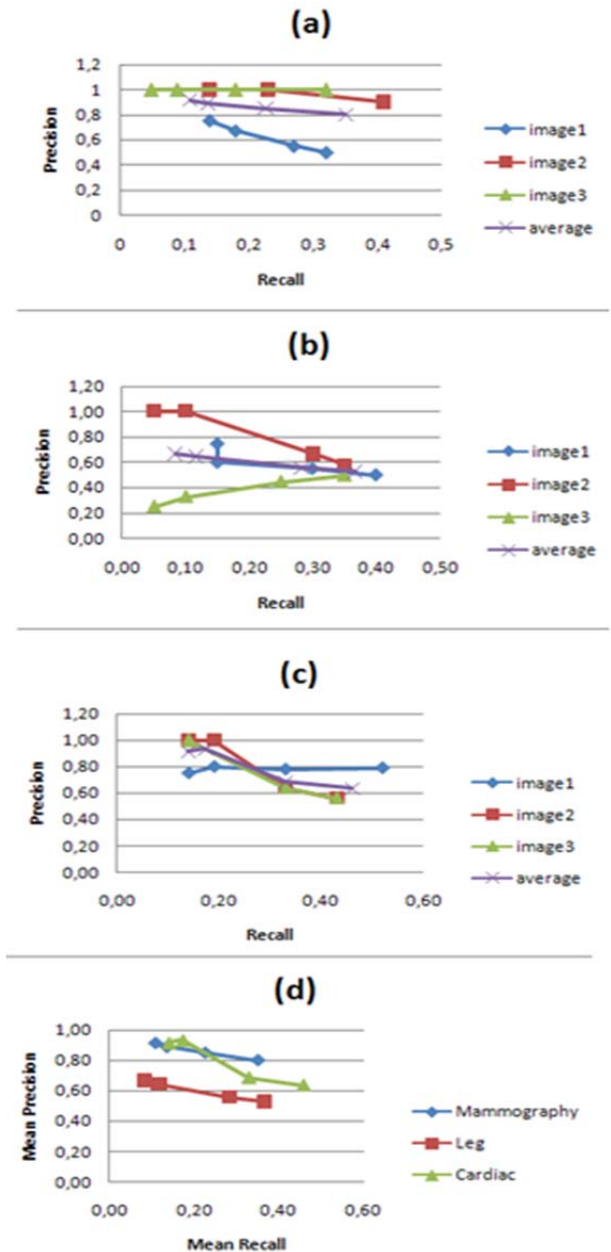

Fig. 7 (a)Precision/recall mammography (b) Precision/recall leg (c) Precision/recall cardiac (d) Mean Precision/recall

Contrary to the images of leg and cardiac, it is apparent that the precision remains high for mammography images; this justifies that the descriptor used to describe the texture is suitable for mammography images.

## V. CONCLUSION

In this paper, we developed a CBIR system of search and image retrieval based on watermarking. The watermark consists of: the descriptor calculated from the pyramidal multi-resolution texture analysis based on LBP, BFILE locator and the patient information.

The watermark is embedded in the image in **offline mode**. Thereby, the image, the descriptor describing its contents, the BFILE locator and patient-related information is a single entity. Therefore, the patient's private information is protected and even inaccessible from the administrator of the database. The chosen watermarking technique is reversible to preserve the contents of the medical image, and provides high data integration capability.